\pdfoutput=1

\documentclass[11pt]{article}

\usepackage{acl}

\usepackage{times}
\usepackage{latexsym}

\usepackage[T1]{fontenc}

\usepackage[utf8]{inputenc}

\usepackage{microtype}

\usepackage{graphicx}
\usepackage{amsmath,amsfonts,tabularx,multirow,makecell}
\title{Improving Conversational Recommendation Systems’ Quality with Context-Aware Item Meta Information}

\author{Bowen Yang$^{\dagger}$,
Cong Han$^{\dagger}$, Yu Li$^{\ddagger}$, Lei Zuo$^{\dagger}$, Zhou Yu$^{\dagger}$, \\
  $^{\dagger}$Columbia University \\
  \texttt{\{by2299, ch3212, lz2771, zy2461\}@columbia.edu} \\
  $^{\ddagger}$University of California, Davis\\
  \texttt{\{yooli\}@ucdavis.edu}}

\begin{document}
\maketitle
\begin{abstract}

\end{abstract}
Conversational recommendation systems (CRS) engage with users by inferring user preferences from dialog history, providing accurate recommendations, and generating appropriate responses. Previous CRSs use knowledge graph (KG) based recommendation modules and integrate KG with language models for response generation. Although KG-based approaches prove effective, two issues remain to be solved. First, KG-based approaches ignore the information in the conversational context but only rely on entity relations and bag of words to recommend items. Second, it requires substantial engineering efforts to maintain KGs that model domain-specific relations, thus leading to less flexibility. In this paper, we propose a simple yet effective architecture comprising a pre-trained language model (PLM) and an item metadata encoder. The encoder learns to map item metadata to embeddings that can reflect the semantic information in the dialog context. The PLM then consumes the semantic-aligned item embeddings together with dialog context to generate high-quality recommendations and responses. Instead of modeling entity relations with KGs, our model reduces engineering complexity by directly converting each item to an embedding. Experimental results on the benchmark dataset \textsc{ReDial} show that our model obtains state-of-the-art results on both recommendation and response generation tasks\footnote{Code is available online \url{https://github.com/by2299/MESE}}.







\section{Introduction}
An automated conversational recommendation system (CRS) \cite{li2019deep, zhou2020improving} is intended to interact with users and provide accurate product recommendations (e.g., movies, songs, and consumables). It has been a focal point of research lately due to its potential applications in the e-commerce industry. Traditional recommendation systems collect user preferences from implicit feedback such as click-through-rate \cite{zhou2018deep} or purchase history and apply collaborative filtering \cite{Su2009ASO, pub.1025106617} or deep learning models \cite{45530, he2017neural} to construct latent spaces for user preferences. Unlike traditional recommendation systems, CRSs directly extract user preferences from live dialog history instead of implicit interactive records, thus can provide better recommendations that precisely address the users' needs. 

Although some progress has been made in this area, there is still room for improvement. First, previous CRSs \cite{chen-etal-2019-towards, zhou2020improving, li2020deux} track entities mentioned in the dialog context, and then search related items in knowledge graphs to recommend to users. However, these systems require a named-entity recognition (NER) module to extract mentioned entities from the dialog context, thus we need to collect additional domain-specific data to train the NER module. In practice, such NER modules have deficient performance, leading to a bad accuracy of CRS. Additionally, entity-dependent recommendation modules lack of contextual information. For example, a user may say: \textbf{\emph{``I watched Ant-Man but did not enjoy it.''}}. In this case, \textbf{\emph{Ant-Man}} is extracted as a mentioned entity and the system will recommend a similar movie to Ant-Man, which is not an appropriate recommendation. Moreover, in some domains, users are more inclined to express their needs with pure language. Since corpora in such domains barely contain named entities, existing entity-based CRSs perform poorly on these domains. Ideally, a CRS should condition its recommendation on the integrated contextual information of the entire dialog context and mentioned entities. 
Second, existing CRSs built upon graph neural networks \cite{kipf2017semisupervised, schlichtkrull2017modeling} cannot quickly scale up or respond to rapid changes of the underlining entities. In e-commerce companies, items for recommendation change daily or even hourly due to constant updates of merchants and products. Existing approaches with graph neural networks require either re-training the entire system when the structure of knowledge graph changes \cite{dettmers2018convolutional} or adding complex architectures on top to be adaptive \cite{wu2019efficiently}. A more flexible architecture can help the system react to rapid changes and adapt itself to new items.

Driven by the motivations above, we present a \textbf{M}etadata \textbf{E}nhanced learning approach via \textbf{S}emantic \textbf{E}xtraction from dialog context i.e. \textbf{MESE}. The major components of MESE contain a pre-trained language model (PLM) and an item encoder architecture. The item encoder takes item metadata as input and is jointly trained with the PLM and dialog context. After training, the item encoder can map item metadata in a systematic way such that contextual information from the dialog is reflected in the constructed embeddings. Item embeddings are then consumed together with dialog context by the self-attention mechanism of the PLM to perform recommendation and response generation. In order for the model to scale up with the size of the item database, we pose recommendation as a two-phase process with candidate selection and candidate ranking following \cite{45530}.


The key contributions of this paper are summarized as follows: This paper presents MESE, a novel CRS framework that considers both item metadata and dialog context for recommendations. Our model employs a simple yet effective item metadata encoder that learns to construct item embeddings during training, thus can adapt to database changes quickly and be independent on task-specific architectures. Extensive experiments on standard dataset \textsc{ReDial} demonstrate that MESE outperforms previous state-of-the-art methods on both response generation and recommendation with a large margin.




\section{Related Work}
Current CRS paradigm contains two major modules: a recommendation module that suggest items based on conversational context and a response generation module that generate responses based on dialog history and the recommended items. How to integrate these two modules to perform well on both tasks has been a major challenge. \cite{chen-etal-2019-towards} leverages external knowledge and employees graph neural networks as the backbone to model entities and entity relations in the knowledge graph (KG) to enhance the performance. In \cite{zhou2020improving}, a word-level KG (ConceptNet \cite{speer2018conceptnet}) is introduced to the system with semantic fusion \cite{sun2020infograph} to enhance the semantic representations of words and items. Since item information and dialog context are processed separately in the above approaches, they suffer from loss of integrated sentence-level information. We propose to condition recommendation on an integrated contextual information of both dialog context and mentioned entities. More recent works try to adopt pre-trained language models (PLM) \cite{vaswani2017attention, Radford2019gpt2, zhang-etal-2020-dialogpt} and template based methods to facilitate response generation. \cite{liang2021learning} generates a response template that contains a mixture of contextual words and slot locations to better incorporate recommended items. \cite{wang2021finetuning} expands the vocabulary list of the PLM to include items to unify the process of item recommendation with response generation. We propose to enhance our PLM with an item metadata encoder to extract context-aware representations by jointly training on both recommendation and response generation tasks. We also generate response templates with slot locations to better incorporate recommended items into responses.

Our work is also inspired by studies from other areas. Recent works have shown that cross-modality training across vision and language tasks can lead to outstanding results in building multimodal representations \cite{tan-bansal-2019-lxmert, lu2019vilbert}. In \cite{tan-bansal-2019-lxmert}, a large-scale transformer based model is adapted with cross-modal encoders to connect visual and linguistic semantics and pre-trained on vision-language pairs to learn intra-modality and cross-modality relationships. Prompt tuning \cite{li-liang-2021-prefix, gao-etal-2021-making} methods prove that PLMs are capable of integrating different sources of information into the same embedding space and perform well on downstream tasks. In terms of using PLM as a recommendation system, \cite{sun2019bert4rec} trains a bidirectional self-attention model to predict masked items and their model outperforms previous sequential models on various recommendation tasks. Inspired by the above studies, we propose to use an encoder module to map item meta information to an embedding space. By jointly training on dialog context and encoded item representations, the system can align these two streams of information by fusing the semantic spaces. The joint embedding representations are then processed by the self-attention mechanism of the PLM to perform recommendation and response generation.

\section{Approach}
In this section, we present our framework MESE that integrates item metadata with dialog context. We first introduce how to encode item metadata and how to process dialog context with the encoded metadata. We then illustrate how the recommendation module and the response generation module are built and how the encoded metadata is incorporated into both modules. Finally, we describe the training objectives and the testing process.

\subsection{Encoding Item Metadata}
\label{sec:encoding}
\begin{figure}[!ht]
    \centering
    \includegraphics[width=1.0\columnwidth]{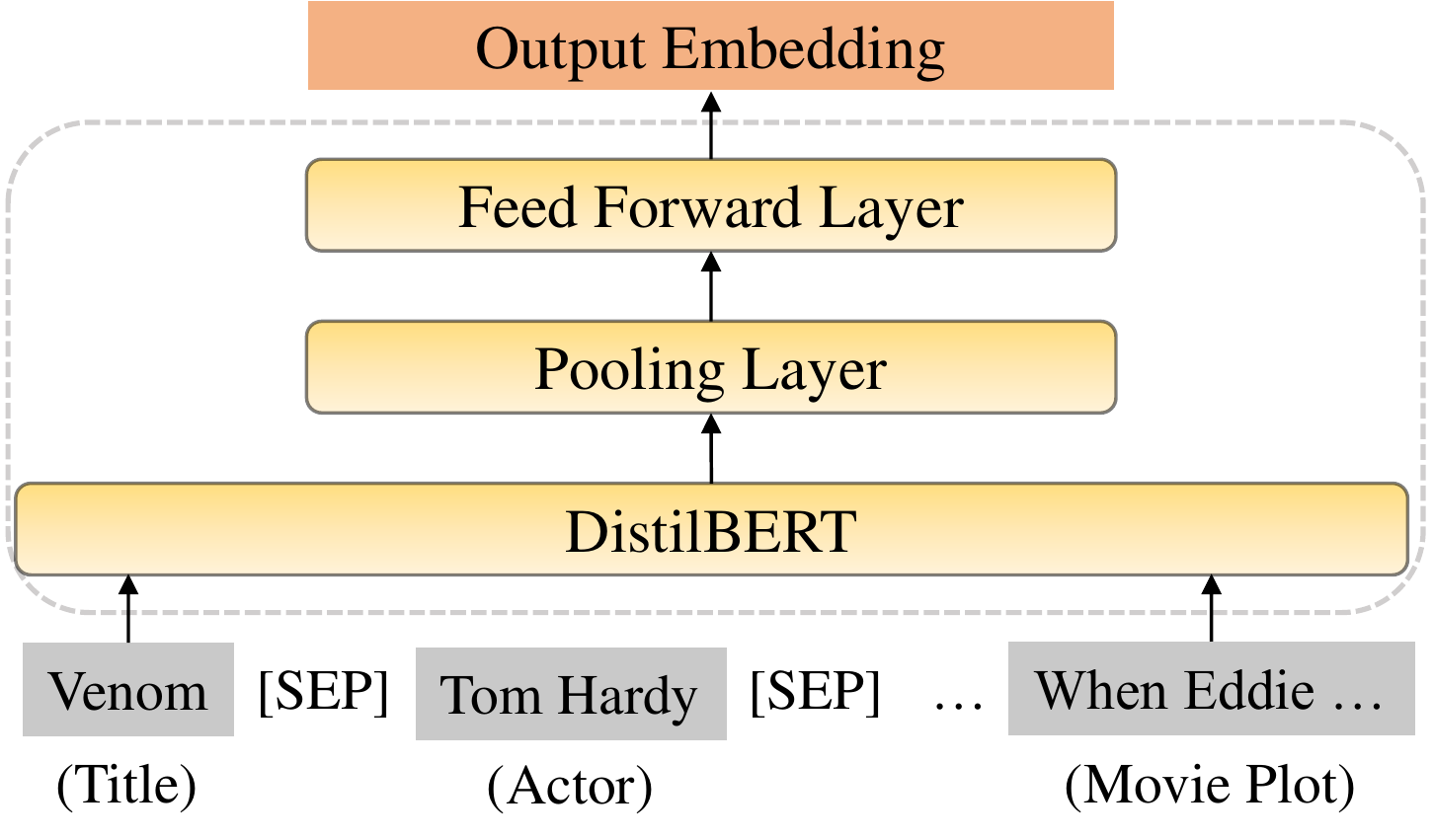}
    \caption{Item Encoder takes in the metadata of an item and outputs an embedding of certain dimensionality.}
    \label{fig:ItemEncoder}
\end{figure}

Instead of modeling item representations based on their relations with other items in the knowledge graphs \cite{chen-etal-2019-towards, wang2021finetuning}, we propose to use an item encoder to directly map the metadata of each item to an embedding. In the movie recommendation setting, description on title, genre, actors, directors, and plot are collected as metadata and concatenated with a "[SEP]" token for each movie. This concatenated information is the input to the item encoder which produces a vector representation for each item. The item encoder consists of a DistilBERT \cite{sanh2020distilbert} model that maps the input sequence to a sequence of vector embeddings, a pooling layer that condenses the sequence embeddings to a single vector embedding, and a feed-forward layer to produce the output embedding. A visualization of this module is shown in Figure \ref{fig:ItemEncoder}. We construct the embeddings for all items in the database.

\begin{figure}[!ht]
    \centering
    \includegraphics[width=1.0\columnwidth]{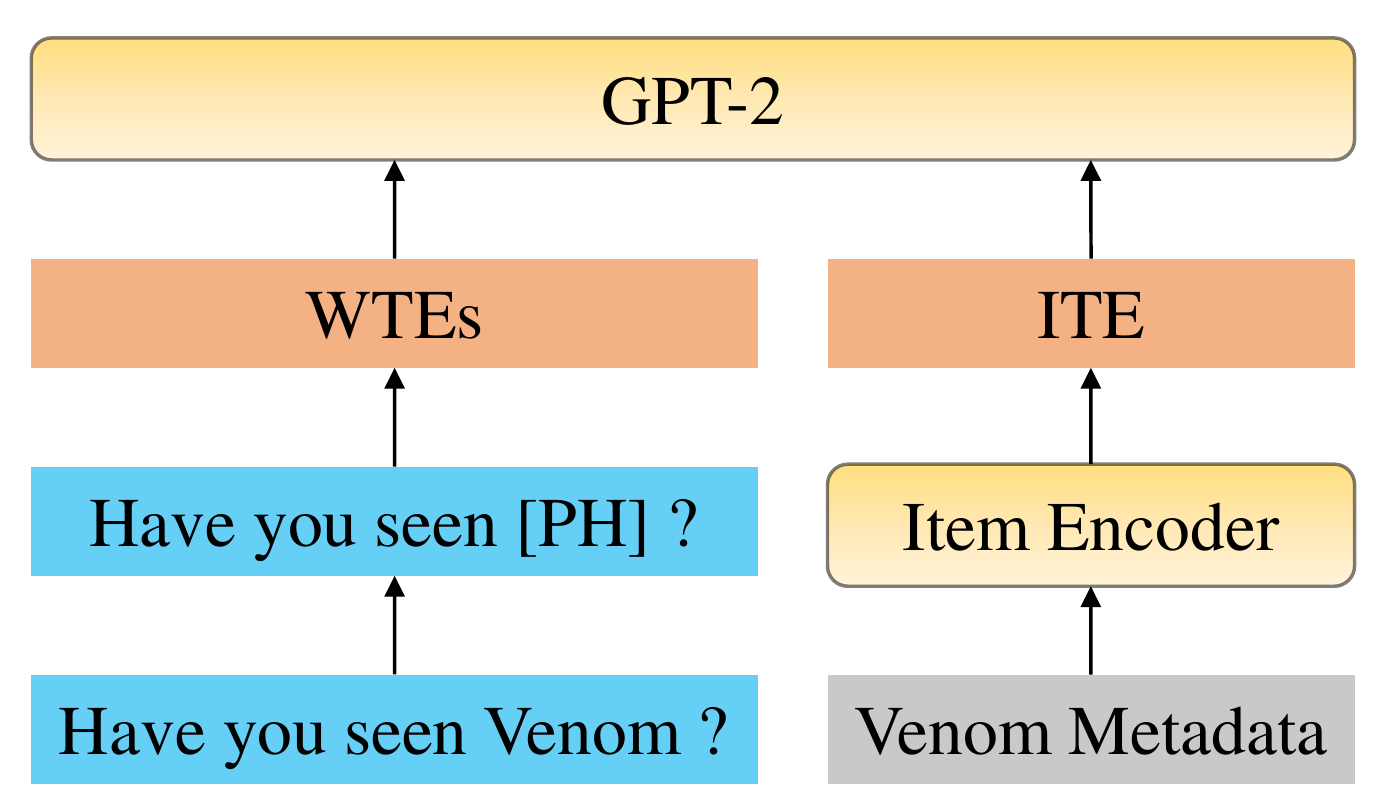}
    \caption{Dialog context is represented as a concatenation of WTEs and ITEs to be consumed by GPT-2.}
    \label{fig:ContextConstruction}
\end{figure}

Next, we discuss how to incorporate items into dialog context with the encoded embeddings and the GPT-2 model \cite{Radford2019gpt2}. Previous studies have shown that KG-based frameworks cannot always integrate recommended items into generated replies \cite{wang2021finetuning}. To solve this issue, we introduce a special placeholder token "[PH]" to the vocabulary list of GPT-2. Every occurrence of item name in the corpus is replaced with this "[PH]" token. This modified dialog sequence is then mapped to a sequence of word token embeddings (WTE) by the vocabulary embedding matrix of GPT-2. An instance of the item encoder is used to encode item metadata into token embeddings. The item encoder takes in item metadata and outputs an item token embedding (ITE) with the same dimensionality as a WTE of the GPT-2 model. The ITE is then concatenated with the WTEs constructed from the dialog context to be consumed by GPT-2. An example is shown in \ref{fig:ContextConstruction}.

\subsection{Recommendation Module}
\label{sec:recommendation}

\begin{figure*}[!t]
    \centering
    \includegraphics[width=2.0\columnwidth]{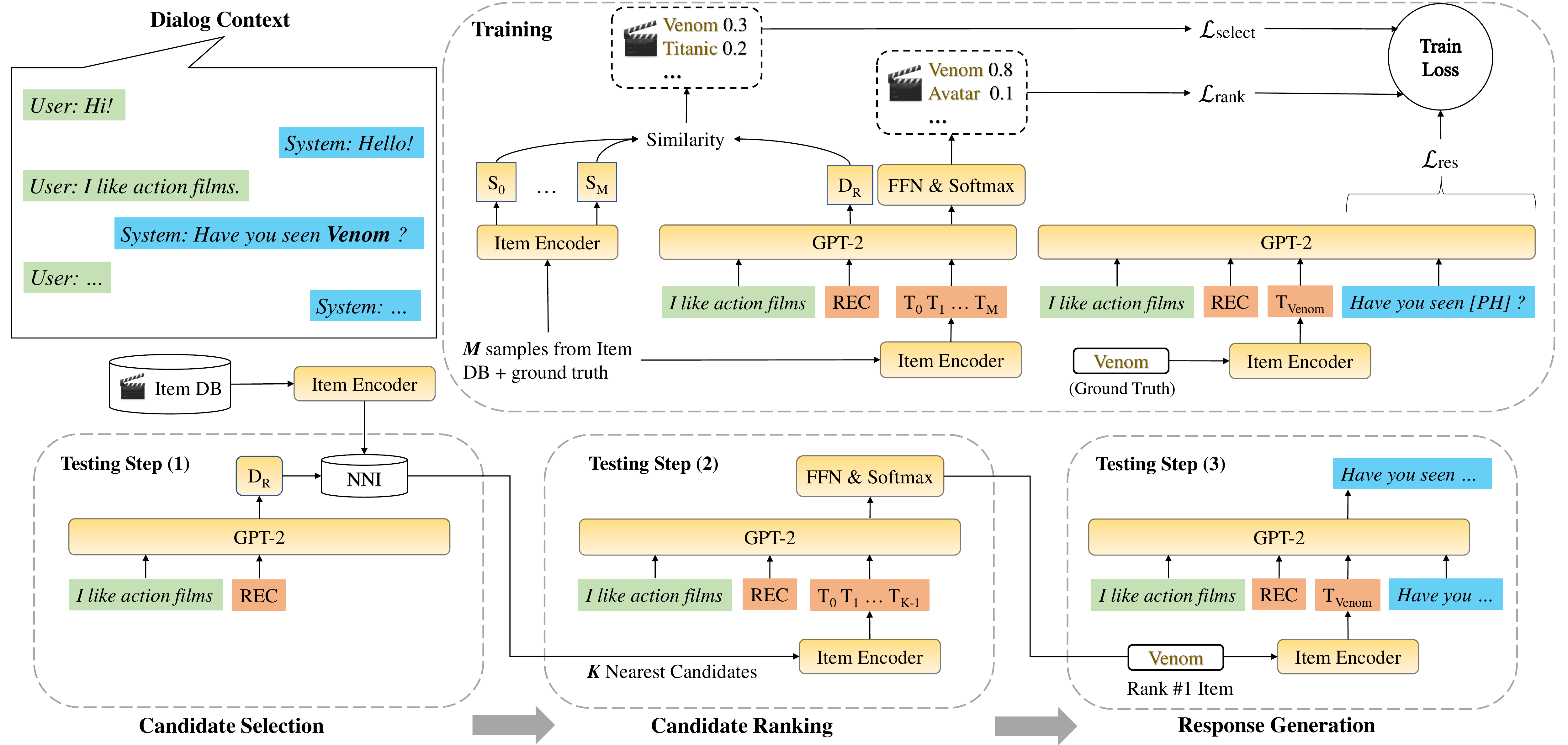}
    \caption{Overview of MESE. During training, M examples are sampled from the item database and participate in computing the joint loss $\mathcal{L}_\text{select}$ and $\mathcal{L}_\text{rank}$, which are then combined with the response generation loss $\mathcal{L}_\text{res}$ and jointly optimized. During testing, the entire metadata DB is stored as a nearest neighbor index (NNI). An approximate nearest neighbor search is performed on $D_{R}$ to get candidate items, which is then fed to the ITE Encoder to compute final scores and the the highest-ranked candidate will be presented to the user in the generated response.}
    \label{fig:model}
\end{figure*}

Similar to \cite{45530}, we pose recommendation as a two-phase process: candidate selection and candidate ranking. During candidate selection, the entire item database is traversed and narrowed down to a few hundred candidates based on a calculated similarity score between the dialog context and the item metadata. During candidate ranking, similarity scores between the dialog context and the generated candidates are calculated with finer granularity because only a few hundred items are being considered rather than the entire database.
The top candidates after sorting is then used as prompts for response generation.

\subsubsection{Candidate Selection}
\label{sec:c_gen}

In this section, we describe the training objective of candidate selection. We add a special token "[REC]" to the vocabulary embedding matrix of GPT-2. This token is used to indicate the start of the recommendation process and to summarize dialog context. At the end of each turn, a token embedding sequence is created following Figure \ref{fig:ContextConstruction} in the format of an interleaving of word token embeddings (WTE) and item token embeddings (ITE) to represent all previous dialog context. When recommendation is labeled in a conversation turn in the training dataset, the WTE of "[REC]" is appended to the previous token embedding sequence to form a new sequence $D$. Next, GPT-2 takes in $D$ and produces an output embedding sequence. We denote the last vector of this output embedding sequence as $D_{R}$ which corresponds to the appended special token "[REC]". $D_{R}$ summarizes dialog context and can be used to retrieve candidate items.

In order to let the model learn how to find candidates based on their relevance to dialog context, we randomly sample M items and their metadata from the database as negative examples and combine them with the ground truth item labeled in the dataset to get the training samples. Another instance of the item encoder, is used to create candidate token embeddings for each item in the training samples. The item Encoder takes in metadata of the samples items and outputs a set of candidate vector embeddings $C = (c_0, c_1, ..., c_M)$, each with the sample dimensionality as $D_{R}$. The recommendation task at this phase is posed as a multi-class classification problem of predicting the ground truth item over the negative samples. The probability of each candidate item is defined in (\ref{form:1}) and optimized by a cross-entropy loss function, denoted as ${\mathcal{L}}_\text{select}$:

\begin{align}
\label{form:1}
P(i) = \frac{e^{c_i \cdot D_{R}}}{\sum_{n=0}^{M} e^{c_n \cdot D_{R}} }
\end{align}

Note that the purpose of this learning objective is to let the model learn how to construct the $D_{R}$ representation instead of learning the probabilities of candidate items. The $D_{R}$ representation is later used in an approximate nearest neighbor search \cite{Liu2004AnIO} to select candidates from the entire database in testing \ref{sec:testing}.

\subsubsection{Candidate Ranking}
\label{sec:c_ranking}

In this section, we describe the training objective of candidate ranking. The goal of candidate ranking is to further perform fine-grained scoring on the similarities between generated candidates and dialog context so that the final rankings of items can better reflect users' preferences. Unlikely previous studies where knowledge graphs are applied to complete this task, we propose to use GPT-2 directly. 

During training, the same context token embedding sequence $D$ and the same training sample with M negative examples are used. The ITE encoder from section \ref{sec:encoding} is used to map the metadata of the sample to an ITE set $T = (t_0, t_1, ..., t_M)$, where the subscript of each $t_i$ corresponds to their index in the database. A concatenation of context sequence $D$ and $T$ are created and consumed by the same GPT-2 model used above and the output embeddings are computed. Note that the order of candidate items should not make a difference on the values of the outputs. Therefore, we add the same positional encoding to each ITE in $T$ and remove the auto-regressive attention masks among the ITEs. We select from the output embeddings the vectors that correspond to the vectors in $T$. For each output vector selected, a feed-forward layer is applied to reduce each vector from a higher dimension to a single number with dimensionality equals 1. This set of numbers are denoted by Q = $(q_0, q_1, ..., q_M)$ where the index of each number corresponds to their index in $T$. The final ranking score of each candidate item is defined in (\ref{form:2}) and optimized by a cross-entropy loss function, denoted as ${\mathcal{L}}_\text{rank}$:

\begin{align}
\label{form:2}
R(i) = \frac{e^{q_i}}{\sum_{n=0}^{M} e^{q_n} }
\end{align}

\subsection{Response Generation Module}
\label{sec:response}

In this section, we describe how to train the model to generate responses based on the recommended items' metadata. The same token embedding sequence $D$ is used as context and current system utterance $U = (w_{0}, w_{1}, ..., w_{n})$ is used as targets where each $w_i$ represents a WTE. We only optimize the GPT-2 model to reconstruct system utterances.

If the current utterance contains recommendations, we create ITEs by passing metadata of the recommended items through the item Encoder used in \ref{fig:ContextConstruction} and append the ITEs to context token embedding sequence $D$ to obtain $D'$. If the current utterance doesn't contain recommendations, $D'$ is simply set to be $D$. The GPT-2 model is trained to reconstruct the ground truth $U$ based on $D'$. The probability of generated response is formulated as:

\begin{align}
P(U | D') = \prod_{i=1}^{n} P(w_{i} | w_{i-1}, ..., w_{0}, D')
\end{align}
The loss function is set to be: 
\begin{align}
\label{loss:language}
L_{res} = -\frac{1}{N} \sum_{i=1}^{N} log( P(U_i | D') )
\end{align}
Where N is the total number of system utterances in one dialog.

\subsection{Joint Learning}
\label{sec:parameter}

Finally, we use the following combined loss to jointly train both the encoders and the GPT-2 model:

\begin{align}
Loss = a \cdot \mathcal{L}_\text{select} + b \cdot  \mathcal{L}_\text{rank} + c \cdot  \mathcal{L}_\text{res}
\end{align}
Where a, b and c are the weights of language training and recommendation training objectives. During training, all weight parameters of the two item encoders, the GPT-2 model and relevant feed-forward layers participate in back-propagation. An overview of training is shown in Figure \ref{fig:model}

\subsection{Testing}
\label{sec:testing}

During testing, a candidate embedding set over the entire item database is built by running metadata through the item encoder used in section \ref{sec:c_gen} and stored with a nearest neighbor index (NNI) \cite{Muja2014ScalableNN}. During response generation, when a "[REC]" token is generated, candidate selection \ref{sec:c_gen} is activated. An approximate nearest neighbor search is conducted over the NNI and K closest candidates are selected based on their similarities from the $D_{R}$ vector. Candidate ranking is then activated and the GPT-2 and the item encoder from \ref{fig:ContextConstruction} are used to generate a score for each candidate. When ranking finishes, the ITE that receives the highest ranking score is appended to the dialog context $D$ and response generation continues until the end of sentence token is generated. After generation is completed, we replace the occurrence of the placeholder token "[PH]" with the title of the recommended item to form the final response. Note that when there is no need for recommendation, our GPT-2 model simply generates a clarification question or a chitchat response with no placeholder tokens. We only present the case when there's only one ground truth recommendation in the utterance. However, it's easy to extend the above approach to multiple recommendations. An overview of testing is shown in Figure \ref{fig:model}

\section{Experiments}
\label{sec:experiments}
In this section, we discuss the datasets used, experiment setup, experiment results on both recommendation and language metrics, and report analysis results with ablation studies.

\subsection{Datasets}
We evaluated our model on two datasets: ReDial dataset \cite{li2019deep} for comparison with previous models and INSPIRED dataset \cite{hayati-etal-2020-inspired} for ablation studies. Both datasets are collected on Amazon Mechanical Turk (AMT) platform where workers make conversations related to movie seeking and recommending following a set of extensive instructions. The statistics of both datasets are shown in Table \ref{table:datasets}

\begin{table}[!ht]
	\centering
	\begin{tabular}{cccc}
		\hline
		Dataset & dialogs & utterances & avg turns \\
		\hline
		ReDial & 10006 & 182150 & 18.2 \\
		INSPIRED & 1001 & 35811 & 10.73 \\
        \hline
    \end{tabular}
	\caption{Statistics of Datasets}
    \label{table:datasets}
\end{table}

\subsection{Experimental Setup}
\subsubsection{baselines}
The baseline models for evaluation on the ReDial dataset is described below:

\textbf{ReDial} \cite{li2019deep}: A dialogue generation model using HRED \cite{sordoni2015hierarchical} as backbone for dialog module

\textbf{KBRD} \cite{chen-etal-2019-towards}: The dialog generation module based on the Transformer architecture \cite{vaswani2017attention}. It exploits external knowledge to perform recommendations and language generation.

\textbf{KGSF} \cite{zhou2020improving}: Concept-net is used alongside knowledge graph to perform semantic-aware recommendations.

\textbf{CR-Walker} \cite{ma2021crwalker}: performs tree-structured reasoning on a knowledge graph and guides language generation with dialog acts

\textbf{CRFR} \cite{zhou-etal-2021-crfr}: conversational context-based reinforcement learning model with multi-hop reasoning on KGs.

\textbf{NTRD} \cite{liang2021learning}: an encoder-decoder model is used to generate a response template with slot locations
to be filled in with recommended items using a sufficient attention mechanism.

\textbf{RID} \cite{wang2021finetuning}: pre-trained language model and knowledge graph are used to improve CRS performance.

\begin{table*}[!t]
	\centering
	
	\begin{tabular}{cccccccccccc}
		\hline
        \multirow{2}{*}{\thead{Model}} &  \multicolumn{4}{c}{Recommendation metrics} &
        \multicolumn{1}{c}{} &\multicolumn{6}{c}{Language generation metrics} \\
        & R@1 & R@10 & R@50 & ReR & & PPL & Dist2 & Dist3 & Dist4 & Bleu2 & Bleu4\\
		\hline
		ReDial & 2.4 & 14.0 & 32.0 & 0.7 & & 28.1 & 0.225 & 0.236 & 0.228 & 0.178 & 0.074\\
		KBRD   & 3.1 & 15.0 & 33.6 & 0.8 & & 17.9 & 0.263 & 0.368 & 0.423 & 0.185 & 0.074 \\ 
		KGSF   & 3.9 & 18.3 & 37.8 & 0.9 & & 5.6 & 0.289 & 0.434 & 0.519 & 0.164 & 0.074 \\
		CR-Walker & 4.0 & 18.7 & 37.6 & - & & - & - & - & - & - & -\\
		CRFR  & 4.0 & 20.2 & 39.9 & - & & - & - & - & - & - & -\\
		RID & - & - & - & 3.1 & & 54.1 & 0.518  & 0.624 & 0.598 & 0.204 & 0.110 \\
		NTRD & - & - & - & 1.8 & & \bf{4.4} & 0.578 & 0.820 & 1.005 & - & - \\
		MESE & \bf{5.6} & \bf{25.6} & \bf{45.5} & \bf{6.4} & & 12.9 & \bf{0.822} & \bf{1.152} & \bf{1.313} & \bf{0.246} & \bf{0.143} \\
        \hline
    \end{tabular}
    \caption{Results and comparison with the literature on \textsc{ReDial}.}
    \label{tab:compare}
\end{table*}

\subsubsection{Implementation Details}
We employed GPT-2 model \cite{Radford2019gpt2} as the backbone of MESE for dialog generation, which contains 12 layers, 768 hidden units, 12 heads, with 117M parameters. We recruited 2 item encoders \cite{sanh2020distilbert} to encoder items in candidate generation \ref{sec:c_gen} and candidate ranking \ref{sec:c_ranking}, respectively, each has a distil-bert model with 6 layers, 768 hidden units, 12 heads, with 66M parameters. We used the AdamW  optimizer \cite{loshchilov2019decoupled} with epsilon set to $1e^{-6}$, learning rate set to $3e^{-5}$. The model was trained for 8 epochs on ReDial dataset, and the first epoch was dedicated to warm up with a linear scheduler. We set the sample size M during candidate generation and candidate ranking to be 150. We chose K = 500 for the number of candidates during testing. We set a=0.3, b = 1.0 and c = 0.5 as coefficients for 3 loss functions respectively.

\subsubsection{Evaluation Metrics}
We performed two evaluations, recommendation evaluation and dialog evaluation, for the model. For recommendation evaluation, we used Recall@X (R@X), which shows whether the top X items recommended by the system include the ground truth item suggested by human recommenders. In particular, we chose R@1, R@10 and R@50 following previous works \cite{chen-etal-2019-towards, zhou2020improving}. We also define recall accuracy of MESE to be the percentage of ground truth items that appear among the 500 generated candidates in the candidate generation phase \ref{sec:c_gen} and ranking accuracy to be the percentage of items that appear in the top k (k=1, 10, 50) position of the sorted candidates in the candidate ranking phase \ref{sec:c_ranking}. The product of the recall and ranking accuracy is the final recommendation accuracy of MESE. We also adopted end-to-end response evaluation following \cite{wang2021finetuning}. We computed response recall (ReR) as whether the final response contains the target items recommended by human annotators. For dialog evaluation, we measured perplexity, distinct n-grams \cite{li-etal-2016-diversity}, and BLEU score \cite{papineni-etal-2002-bleu}.

\section{Experimental Results}
\subsection{Model Results}

We first report recall, ranking, and final accuracy on \textsc{ReDial} dataset of MESE in table \ref{tab:RRF}. From the results, it can be seen that candidate ranking has remarkable performance gains in scoring the items. It demonstrates that pre-trained language model have great potential in making recommendations. One possible reason behind this is that the self-attention mechanism is effective in learning the discrepancies between item semantics and dialog semantics.

\begin{table}[!ht]
	\centering
	\begin{tabular}{cccc}
		\hline
		top k & Ranking Acc & Recall Acc & Final Acc  \\
		\hline
		@1      & 7.2 & 0.778 & \bf{5.6} \\
		@10     & 33.0 & 0.778 & \bf{25.6} \\ 
		@50     & 58.5 & 0.778 & \bf{45.5} \\
        \hline
    \end{tabular}
	\caption{Recall, Ranking and Final Accuracy of MESE.}
    \label{tab:RRF}
\end{table}

Table \ref{tab:compare} compares different models on \textsc{ReDial} dataset. The superiority of MESE persists across recommendation and language generation. On all recommendation metrics, including R@1, R@10, and R@50, MESE outperforms the state-of-the-art models by a large margin. We argue in \ref{sec:ablation} that this significant gain of performance is due to the effectiveness of the item encoder. MESE also performs well on the ReR score, which indicates that the filling placeholder tokens can help integrate recommended items into responses. For language generation, MESE also achieves significantly better performance than all other models on distinct ngrams and bleu scores with the exception that the PPL is worse than those of KGSF and NTRD. This indicates that MESE can generate more diverse responses while sticking to the topic.

\subsection{Ablation Studies and Analysis}
\label{sec:ablation}

In this section, we first analyze the reason behind the performance gain of our recommendation module by analyzing the embeddings learned by the item encoder.

\textbf{How much does metadata help with recommendation?} We argue that our training objectives on recommendation enable the item encoder to selectively extract useful features pertinent to the recommendation task from item metadata and construct item representations that resonate with instructional semantic properties in the dialog histories. For example, in \textsc{ReDial} dataset, movie genre information is the most frequently mentioned property in dialog histories and human recommenders often make recommendation decisions based on this property. Although other properties like actors also help with recommendation, they do not appear in the corpus as often as genres or movie plots. We designed the following experiments to test our hypothesis. First, we train MESE with movie genre and plot information removed from the metadata, which we refer to as MESE w/o content, and compare its recommendation performance with MESE in Table \ref{tab:MESEx}.

\begin{table}[!ht]
	\centering
	\begin{tabular}{cccc}
		\hline
		Model & R@1 & R@10 & R@50 \\
		\hline
		MESE w/o content     & 3.9 & 19.5 & 37.9 \\
		MESE       & \bf{5.6} & \bf{25.6} & \bf{45.5} \\
        \hline
    \end{tabular}
	\caption{Comparison Results of MESE and MESE w/o content.}
    \label{tab:MESEx}
\end{table}

As we can see from the table, there is a significant performance decrease after we remove genre and plot information, which indicates that MESE depends on the input features to make high quality recommendations. We also point out that movie titles contain weak genre information but are not able to provide adequate features for the item encoder to extract from.

\textbf{How does the item encoder help with recommendation?} Specifically, we select all movie items with only one genre as our candidates, resulting in a subset of \textasciitilde 600 movies. We then select 2 item encoders (section \ref{sec:c_ranking}) from MESE, MESE w/o content, and the item encoder before training (MESE raw), respectively, and obtain 3 sets of item embeddings by passing the movie subset to these encoders. On each set of embeddings, we run a K-means clustering algorithm with K being set to be 3, 4, and 5, respectively. For each cluster obtained, we calculated the proportion of the majority genre among all item candidates. This process is repeated 20 times for each K and the average accuracy is reported in Table~\ref{tab:embeddingAnalysis}. The reasoning behind this setup is that the more features a set of embeddings contain about genre, the closer the clusters should be towards its central point, thus the higher the accuracy should be for its majority genre.

\begin{table}[!ht]
	\centering
	\begin{tabular}{cccc}
		\hline
		Model & K=3 & K=4 & K=5 \\
		\hline
		MESE raw      & 0.492 & 0.514 & 0.574 \\
		MESE w/o content     & 0.555 & 0.589 & 0.606 \\
		MESE       & \bf{0.695} & \bf{0.725} & \bf{0.738} \\
        \hline
    \end{tabular}
	\caption{Item Encoders Clustering Accuracy}
    \label{tab:embeddingAnalysis}
\end{table}

As we can see from the table, without training, MESE raw, being the least sensitive to genre information, achieves the lowest accuracy scores on all clusters. MESE w/o content, although deprived of genre and plot, still has slightly higher accuracy than MESE raw due to its exposure to \textsc{ReDial} conversations. MESE is most sensitive to genre information. This is an indication that by aligning recommendation related information in both dialog context and item metadata, our model is able to generate meaningful representations for the task, which can facilitate the language model to produce better rankings through its multi-head attention mechanism and result in better recommendation performance. Previous KG-based recommendation module, although also modeled metadata as relations with Graph Convolutional Networks, did not compose item representations in a systematic way.

\textbf{What if we remove mentioned entities from dialog context?} Mentioned entities are crucial to previous approaches \cite{chen-etal-2019-towards, zhou2020improving} in terms of recommendations. We train MESE with mentioned entities removed from dialog history and compare its performance with MESE on \textsc{ReDial} dataset and INSPIRED dataset in table \ref{tab:no_entites}.

\begin{table}[!ht]
    \small
	\centering
	\begin{tabular}{ccccc}
		\hline
		Dataset & Model & R@1 & R@10 & R@50 \\
		\hline
		\textsc{ReDial} & MESE w/o item & 3.4 & 18.1& 38.7 \\
		& MESE & \bf{5.6} & \bf{25.6} & \bf{45.5} \\
		\hline
		INSPIRED & MESE w/o item & 4.3 & 11.9 & 26.7 \\
		  & MESE  & \bf{4.8} & \bf{13.5} & \bf{30.1} \\
        \hline
    \end{tabular}
	\caption{Results of MESE and MESE w/o on \textsc{ReDial} and INSPIRED.}
    \label{tab:no_entites}
\end{table}

We can see removing the entities led to an average of 26.3\% performance drop on \textsc{ReDial} and an average of 11.2\% performance drop on INSPIRED. The recommendation performance on \textsc{ReDial} are more impacted by the removal of entities. That is because the conversations in \textsc{ReDial} are rich with entities and weak in semantic information, whereas, INSPIRED is more sparse on entities but contains more semantics. \textsc{ReDial} has 10006 conversations, in which there is 1 mentioned movies among every 21.85 word tokens. Its sentence level distinct 1-grams and 3-grams are 0.15 and 2.81. However, INSPIRED dataset has 1001 conversations, in which there is 1 mentioned movies among every 63.54 word tokens. Its sentence level distinct 1-grams and 3-grams are 0.59 and 6.84. This proves that our model can efficiently infer user interests from texts to make high-quality recommendations without explicitly using mentioned entities. This property could be useful in an e-commerce setting where users tend to convey their requirements more with texts than entities. It could also be useful in a cold start scenario where we don't have many entities in the context and are forced to only rely on semantics.

\section{Conclusion and Future Work}
In this paper, we introduced MESE, a novel CRS framework. By utilizing item encoders to construct embeddings from metadata, MESE can provide high-quality recommendations that align with the dialog history. We also analyzed various behaviors of MESE to better understand its underlining mechanisms. Our approach yields better performance than existing state-of-the-art models. As for future work, we will consider applying this approach to a broader domain of CRS datasets. Currently, we only experiment on movie recommendations. As mentioned in section \ref{sec:ablation}, MESE is capable of efficiently utilizing dialog history as a whole and construct item embeddings that reflects user preferences. It has the potential to work well with cross-modal tasks. For example, multimodal works can be explored in the e-commerce domain with MESE based architectures.

\bibliography{anthology,custom}
\bibliographystyle{acl_natbib}


\end{document}